\pdfoutput=1

\documentclass[11pt]{article}

\usepackage{acl}

\usepackage{times}
\usepackage{latexsym}

\usepackage[T1]{fontenc}

\usepackage{bm}
\usepackage{bbm}
\usepackage{amssymb} 

\usepackage[utf8]{inputenc}

\usepackage{microtype}

\usepackage{amsmath}
\usepackage{graphicx}
\usepackage{float}
\usepackage{subcaption}
\usepackage{amsfonts}
\usepackage{multicol}
\usepackage[bottom]{footmisc}
\usepackage{booktabs}
\usepackage{multirow}

\usepackage{pifont}
\newcommand{\cmark}{\ding{51}}%
\newcommand{\xmark}{\ding{55}}%
\newcommand{\br}[1]{\bm{\mathrm{#1}}}
\newcommand{\std}[1]{\scriptsize{$\pm$#1}}
\newcommand{\argmax}[1]{\underset{#1}{\text{argmax}}\;}
\newcommand{\argmin}[1]{\underset{#1}{\text{argmin}}\;}

\title{Mitigating Word Bias in Zero-shot Prompt-based Classifiers}

\author{Adian Liusie, Potsawee Manakul, Mark J. F. Gales \\
  ALTA Institute, Department of Engineering, University of Cambridge \\
  \texttt{al826@cam.ac.uk, pm574@cam.ac.uk, mjfg@eng.cam.ac.uk}}

\begin{document}
\pagenumbering{arabic}
\pagestyle{plain}
\maketitle
\begin{abstract}

    Prompt-based classifiers are an attractive approach for zero-shot classification. However, the precise choice of the prompt template and label words can largely influence performance, with semantically equivalent settings often showing notable performance difference. This discrepancy can be partly attributed to word biases, where the classifier may be biased towards classes. To address this problem, it is possible to optimise classification thresholds on a labelled data set, however, this mitigates some of the advantages of prompt-based classifiers. This paper instead approaches this problem by examining the expected marginal probabilities of the classes. Here, probabilities are reweighted to have a uniform prior over classes, in an unsupervised fashion. Further, we draw a theoretical connection between the class priors and the language models' word prior, and offer the ability to set a threshold in a zero-resource fashion. We show that matching class priors correlates strongly with the oracle upper bound performance and demonstrate large consistent performance gains for prompt settings over a range of NLP tasks.\footnote{code available on github at \url{https://github.com/adianliusie/robust-prompt-classifier}}
\end{abstract}

\section{Introduction}
Large language models (LLM) have shown impressive general ability for natural language processing (NLP) tasks. LLMs can effectively handle a range of NLP tasks through ‘prompting’, where a natural language instruction is added to the input, conditioning the model to the task at hand. Prompting can either be an emergent ability learned through scaling up model size \cite{brown2020language, wei2022emergent} or an ability learned through instruction tuning \cite{weifinetuned, chung2022scaling, ouyang2022training}. Despite the recent popularity of prompting, there is a known sensitivity of prompt-based LLMs to elements such as prompt template and label words \cite{gao2021making, schick2021exploiting}. Previous works have demonstrated that prompt templates can significantly impact task performance \citep{shin2020autoprompt, zhou2022large} and that factors such as chosen label words can influence system performance for classification tasks \citep{zhao2021calibrate, holtzman2021surface}. 

\begin{figure}[!t]
    \centering
    \includegraphics[width=\columnwidth]{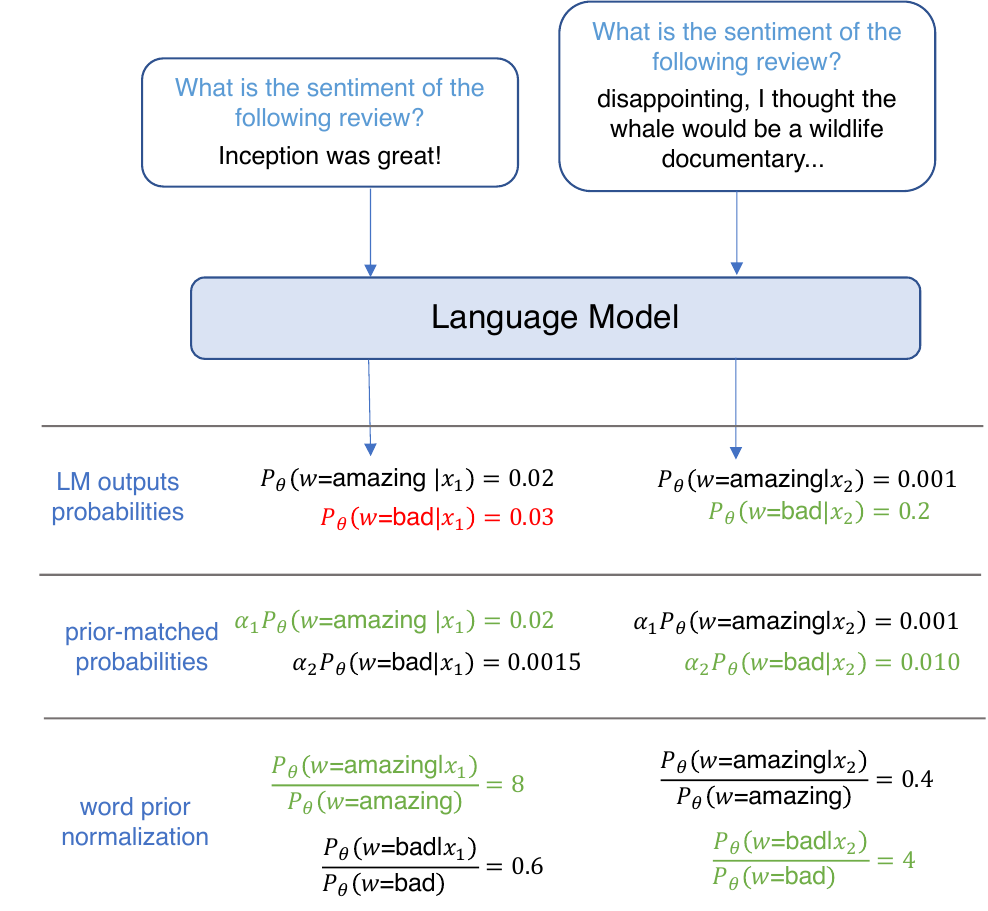}
    \caption{Instead of using the raw LM output probabilities of the label words, we consider mitigating bias by finding weights that make the classifier unbiased over classes. This is connected to normalising by word priors, which we use as a zero-resource de-biasing approach.}
    \label{fig:my_label}
\end{figure}

This work focuses on the influence of `word biases' for prompt-based classifiers. i.e. the bias that prompts may have towards certain classes, independent of the input text. To account for this bias, one could use a labelled dataset to find optimal class decision thresholds. This, however, requires labelled task data, which may limit the zero-shot benefits of prompt-based classifiers. We propose a simple unsupervised solution of re-weighting probabilities, where we use unlabelled data to search for weight parameters that ensure a uniform prior over classes. We show that this prior matching leads to greater robustness for diverse prompt settings and that the unsupervised weights which debias the classifier is highly correlated with the oracle weights that maximise accuracy. Further, we provide theoretical analysis that draws a connection between word priors and inherent class bias, which we use to motivate a zero-resource normalisation approach that is competitive with prior matching. Overall, we demonstrate that our unsupervised approach highly reduces sensitivity to the chosen prompt and label words, and that settings which initially fail can often be made effective through a simple probability re-weighting.

Our contributions are 1) We propose a simple unsupervised probability re-weighting method, and empirically demonstrate greater robustness to prompt and label word choice, with large accuracy gains across prompt settings for a range of standard NLP tasks. 2) We theoretically connect the weight parameters to word priors and use this to motivate a zero-resource re-weighting approach. 3) We show that the weights of prior matching are highly correlated with the optimal oracle weights that maximize accuracy, illustrating that our approach is a near-optimal use of a system's output probabilities. 

\section{Mitigiating Bias by Re-weighting}
\textbf{Prompt-based classifiers} Given an input sequence $\br{x}\!\in\!\mathcal{X}$, large language models (LLMs) model $P_{\theta}(\br{w}|\br{x})$, the output probability distribution over all possible sequences $\br{w}\!\in\!\mathcal{X}$. For a classification task $\mathcal{T}$, a prompt-based classifier 1) reformats the input text $x$ to prompt $\br{p}\in\mathcal{X}$ by including the task instruction, and 2) selects class words $\{w_i\}_{1:K}$ which are associated to each output class $\{y_i\}_{1:K}$. For example in sentiment classification, one can use prompt \textit{`what is the sentiment of the following review? <x>'}, (where <x> is the current input $x$, e.g. \textit{`Inception was absolutely brilliant'}), and class words $w_0$=\textit{bad} and  $w_1$=\textit{good} for the negative and positive classes respectively. For a prompt classifier, $Q=\{\br{p}, \{w_i\}_{1:K} \}$, class probabilities can be set to be proportional to the probability of the associated class word, where the final decision $\hat{y}$ is the class with the highest probability \cite{zhao2021calibrate, jiang2020can}.
\begin{align}
    \tilde{P}_{\theta}(y_k| \br{x}, Q) &= \frac{P_{\theta}(w_k|\br{p}(\br{x}))}{\sum_{w_i} P_{\theta}(w_i|\br{p}(\br{x})) } \label{eq:baseline} \\
    \hat{y} &= \underset{k}{\text{argmax}} \; \tilde{P}_{\theta}(y_k|\br{x}, Q)
    \label{eq:argmax}
\end{align}

\noindent However, as a large language model, the prompt-based classifier may return probabilities that are influenced by distributional statistics of words \cite{gardner2021competency, liusie-etal-2022-analyzing}. This may lead to inherent class bias, where label words may have high probability not because they better answer the prompt, but because they have a high LM prior. 

\vspace{2mm}
\noindent\textbf{Optimal Weights} To account for this, one can define weight parameters $\bm{\alpha} = \{\alpha_i \}_{1:K}$, where each $\alpha_i \in \mathbb{R}^+$ scales the probabilities of the classifier, 
\begin{align}
    \hat{P}_{\theta}(y_k| \br{x}, Q, \bm{\alpha}) = \frac{\alpha_k \tilde{P}_{\theta}(y_k| \br{x}, Q)}{\sum_i \alpha_i \tilde{P}_{\theta}(y_i| \br{x}, Q)}
\end{align}

\noindent Given labelled task dataset $\mathcal{D}=\{(\br{x}^{(j)}, y^{(j)})\}_{j=1}^N$, one can then find the optimal weights $\bm{\alpha}^*$ that maximises the accuracy of the prompt classifier $\hat{P}_{\theta}(y_k| \br{x}, Q, \bm{\alpha})$ over the dataset, 
\begin{align}
    \bm{\alpha}^* &= \argmax{\bm{\alpha}} \text{Accuracy}(Q, \bm{\alpha}, \mathcal{D}) \label{eq:optimal}
\end{align}

\begin{table*}[h!]
    \centering
    \begin{tabular}{ccc|cccccc}
        \toprule
        method  & inputs & labels & imdb & rt & amazon & snli & mnli & qqp \\
        \midrule
        baseline       & \xmark & \xmark & 85.4\std{12.7} & 78.8\std{14.0} & 86.0\std{13.8} & 45.2\std{13.7} & 43.5\std{11.3} & 65.4\std{14.0} \\ 
        null-input & \xmark & \xmark & 92.1\std{3.2}  & 89.1\std{3.8}  & 95.0\std{1.8}  & 75.2\std{10.4} & 66.1\std{9.7} & 77.4\std{6.6} \\
        prior-match & \cmark & \xmark & 93.1\std{3.3} & 90.9\std{1.6} & 96.0\std{0.8} & 78.5\std{9.3} & 69.8\std{9.7} & 79.1\std{2.4}\\
        optimal& \cmark & \cmark & 93.5\std{2.7} & 91.2\std{1.5}   & 96.1\std{0.7} & 79.4\std{8.2} & 70.8\std{8.6} & 82.3\std{2.8}\\
        \bottomrule
    \end{tabular}
    \caption{Average dataset accuracy and standard deviations, over all prompts and label words. \textbf{baseline} and \textbf{null-input} are zero-resource classification methods, \textbf{prior matching} uses the text inputs but not labels, while \textbf{optimal} is an oracle approach that uses the labels to search for the best thresholds. Results for FlanT5 large}
    \label{tab:task_table}
\end{table*}

\noindent \noindent\textbf{Prior-Matching} The previous approach requires labelled data, which may limit the benefit of using prompt-based classifiers. As an alternative, one can find the values $\bar{\bm{\alpha}}$ that ensure that the classifier is unbiased, such that the class prior $\hat{P}(y_k|Q, \bm{\alpha})$ matches the true prior $P(y_k)$
\begin{align}
    \hat{P}_{\theta}(y_k|Q, \bm{\alpha}) &= \mathbb{E}_{\br{x}} \{ \hat{P}_{\theta}(y_k|\br{x}, Q, \bm{\alpha})\} \\
    &\approx \frac{1}{N} \sum_{j=1}^N  \hat{P}_{\theta}(y_k^{(j)}|\br{x}^{(j)}, Q, \bm{\alpha})
\end{align}
\begin{equation}
    \bar{\bm{\alpha}} = \argmin{\bm{\alpha}} \sum_{\forall y_k} | \hat{P}_{\theta}(y_k|Q, \bm{\alpha}) - P(y_k)| \label{eq:unbiased}
\end{equation}

\noindent A deterministic solution that exactly matches the distributions exists, which can be found with a search with 1 degree of freedom (that can be accounted for by setting $\alpha_1 = 1$). If there is no expected class bias, one can assume equal probabilities over all classes, $P(y_k) = \mathcal{U}(y_k) = \frac{1}{N}$. This approach is therefore unsupervised and only requires text inputs $\mathcal{D}_{x}=\{\br{x}^{(j)}\}_{j=1}^M$, which therefore can be applied at inference to any test set. \\

\noindent \textbf{Null-Input Approximation} The dependence of prior-matching on unlabelled dataset $\mathcal{D}_{x}$ is a drawback. In Appendix \ref{sec:proof_1}, we show that one can make the analytical approximation
\begin{equation}
    \bar{\alpha}_k \approx \frac{1}{\mathbb{E}_{x}\{ P_{\theta}(w_k|\br{x}, Q) \}} = \frac{1}{P_{\theta}(w_k| Q)} \label{eq:monte-norm}
\end{equation}

\noindent Inspired by \citet{zhao2021calibrate}, we consider a resource-free approximation of the word prior (equation \ref{eq:monte-norm}) by considering the output word probabilities of the null input $\emptyset$ (i.e. an empty string). 
\begin{align}
    P_{\theta}(w_k | Q)  &\approx  P_{\theta}(w_k|\br{p}(\emptyset)) \label{eq:null_norm}
\end{align}

\noindent This enables a zero-resource approximation of weight parameters $\bar{\bm{\alpha}}$.

\section{Experiments}
\subsection{Experimental Setup}
\noindent\textbf{Data} 
Experimental results are run on standard NLP benchmarks, including sentiment classification (\textit{IMDB} \cite{maas-etal-2011-learning}, \textit{Rotten Tomatoes} \cite{pang-lee-2005-seeing} and \textit{amazon}), natural language inference (\textit{SNLI} \cite{bowman-etal-2015-large} and \textit{MNLI} \cite{williams-etal-2018-broad}) and paraphrase detection (\textit{QQP} \cite{wang-etal-2018-glue}). Evaluation is reported on standard test sets, except for amazon polarity where 5000 test examples were randomly sampled. \\

\noindent\textbf{Models} We use FlanT5 large\footnote{https://huggingface.co/google/flan-t5-large} \cite{chung2022scaling}, a T5 with further instruction-tuning stage where the system was trained in a multi-task fashion over 1,836 tasks, each prepended with a natural instruction prompt. This work evaluates FlanT5 for different NLP tasks with arbitrary prompting set ups. For each task, we select 6 prompt templates and for binary classification tasks consider 25 possible class words pairs, while for NLI we have 64 class word triplets (where all permutations of valid class words are considered). All prompts and label words used are given in Appendix \ref{sec:example_prompts}. Further experiments for FlanT5 base and Llama-2-chat can be found in Appendix \ref{sec:app_tables}. \\

\noindent\textbf{Methods} We consider 4 different methods to leverage LLM probabilities for classification. Class word probability via equation \ref{eq:baseline} (\textbf{baseline}). Normalised probabilities calculated using null-inputs priors via equation \ref{eq:null_norm} (\textbf{null-input}). Optimising $\alpha_k$ with a search to have unbiased class prior via equation \ref{eq:unbiased} (\textbf{prior-match}). The oracle upper-bound performance, found by optimising the optimal accuracy threshold via equation \ref{eq:optimal} (\textbf{optimal}). 

\subsection{Experimental Results}
\begin{figure*}[ht!]
    \centering
    \includegraphics[width=\linewidth]{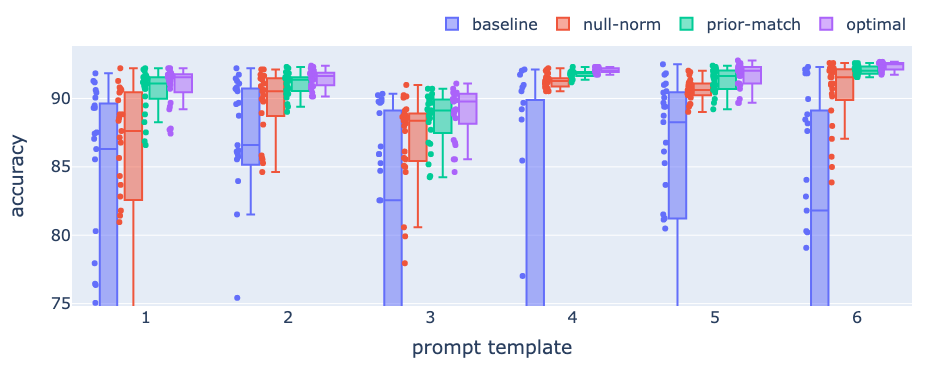}
    \caption{boxplots of the accuracy of all label-word pairs for \textbf{rotten tomatoes}, over all the considered prompts}
    \label{fig:boxplot_rt}
\end{figure*}


\noindent\textbf{Classification Robustness} Table \ref{tab:task_table} shows the mean and standard deviation of accuracies among all prompt and class word settings for a given task. We observe large consistent gains from both re-weighting approaches, with prior-matching increasing baseline accuracy by between 6.7\% to 12.1\% for sentiment classification, 13.7\% for qqp, and over 25\% for natural language inference. Prior-matching also demonstrates performance very similar to the oracle upper-bound, often within 1\%, showing that the unsupervised prior-match approach is competitive with the supervised threshold search. Prior-matching also performs better than null-input by a small margin in all tasks, where this small gap confirms that the word-prior normalisation is a very reasonable zero-shot approximation. \\

\noindent\textbf{Prompt Robustness} Figure \ref{fig:boxplot_rt} illustrates a boxplot of rotten tomatoes performance over all class-words for each considered method, over all 6 prompts. As observed in Table \ref{tab:task_table}, naively using raw label word probabilities (dark blue) leads to considerable fluctuations in accuracy; some prompt and label word settings lead to reasonable accuracy (92\%+ accuracy), however there is observed brittleness to label word choice, with many settings demonstrating poor performance. Prior matching (green) leads to significant robustness, with nearly all sensible settings above 85\% accuracy. We further find that, as shown in Table \ref{tab:task_table}, the unsupervised approach has accuracies very comparable to those when using optimal thresholds. 

\begin{figure}[h!]
    \centering
    \includegraphics[width=\columnwidth]{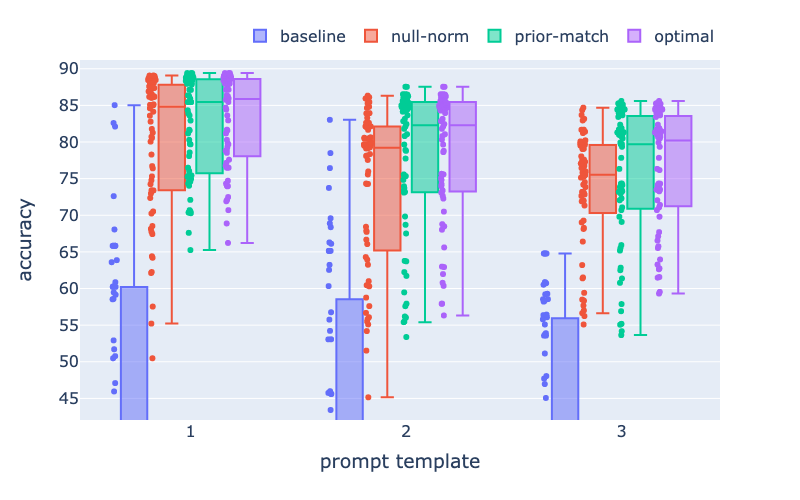}
    \caption{boxplots of the accuracy of all label-word sets for \textbf{snli}, for the first 3 prompts}
    \label{fig:boxplot_snli}
\end{figure}

In Figure \ref{fig:boxplot_snli}, we consider similar boxplots for SNLI and observe larger gains through reweighting. This was as higher probabilities are often assigned to the entailment and contradiction labels words, leading to under-classification of the neutral class. We observe greater sensitivity to prompt choice and label words for snli than as observed in rotten tomatoes, even with reweighting. \\

\begin{figure}[h!]
    \centering
    \includegraphics[width=\columnwidth]{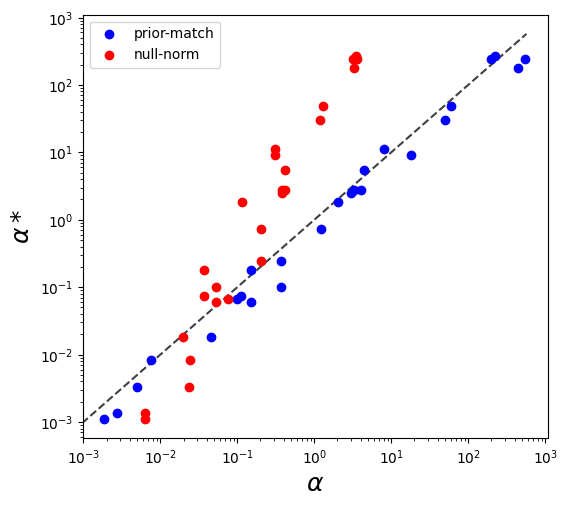}
    \caption{Scatter plot of the optimal weights $\bm{\alpha}^*$ (equation \ref{eq:optimal}) with the prior match weights $\bm{\bar{\alpha}}$ (equation \ref{eq:unbiased}) and the approximation via null-input (equation \ref{eq:null_norm}), for all settings of prompt 1 on \textbf{amazon}}
    \label{fig:threshold_lineplot}
\end{figure}

\noindent \textbf{Weight Alignment} Figure \ref{fig:threshold_lineplot} shows a scatter plot of the weights found by the optimal threshold search $\bm{\alpha}^*$ (equation \ref{eq:optimal}), with those found from the unsupervised prior matching method $\bar{\bm{\alpha}}$ (equation \ref{eq:unbiased}) and the zero-resource word prior approximation (equation \ref{eq:null_norm}). We see a clear linear relationship between optimal and prior-match, illustrating that accounting for the marginal bias is almost equivalent with maximising accuracy, however, achieved in an unsupervised fashion. Null-input is also well correlated with the optimal thresholds, but there is a less direct relationship. Similar linear relationships are observed also for other binary-classification tasks and prompts, as shown in Appendix \ref{sec:linear_plots}. \\

\section{Conclusions}
This paper analyzes prompt-based classifiers and demonstrates that inherent class bias is a significant factor that influences the sensitivity of the system to prompt and label words. We propose an unsupervised approach of prior matching, which we demonstrate performs competitively to the supervised alternative of searching for optimal thresholds, while avoiding the need for labelled data. We relate prior matching with word biases, and motivate a zero-resource approach of debiasing model probabilities. We show that our methods lead to practical approaches that reduce the sensitivity to design choices such as prompts and label words.

\section*{Limitations}
This work considered sentiment classification, natural language inference, and paraphrase detection, and could have been extended over a greater suite of tasks to guarantee its effectiveness. Further, this paper ran experiments on FlanT5 and Llama2, and this work has not yet explored a larger range of prompted language models. FlanT5 has also been instruction-tuned on similar tasks, so the findings may be limited in scenarios where known capabilities have to be elicited from models robustly. 

\section*{Ethical Considerations}
Though this work suggests methods to improve the robustness of prompt-based classifiers to prompts and label words, this does not imply that all design choices will work. In some set ups, the system may be ineffective and have poor generalisation over the task. Deploying machine learning classifiers in real-world classification settings has many associated risks, and careful analysis should be made before deploying such systems. 

\section*{Acknowledgements}
This work is supported by Cambridge University Press \& Assessment (CUP\&A), a department of The Chancellor, Masters, and Scholars of the University of Cambridge, and the Cambridge Commonwealth, European \& International Trust.

\bibliography{anthology,custom}
\bibliographystyle{acl_natbib}

\appendix
\section{Derivation of Zero-Resource Equation}
\label{sec:proof_1}
For a prompt classifier $Q=\{\br{p}, \{w_i\}_{1:K} \}$, class probabilities are assumed to be proportional to the probability of the associated class word,
\begin{align}
    \tilde{P}_{\theta}(y_k| \br{x}, Q) &= \frac{P_{\theta}(w_k|\br{p}(\br{x}))}{\sum_{w_i} P_{\theta}(w_i|\br{p}(\br{x}))}
\end{align}

\noindent Given the task dataset $\mathcal{D} = \{ \{ \br{x}^{(j)}, y^{(j)} \} \}_{j=1}^N$, one can calculate the assumed prior of the prompt classifier over the output classes,
\begin{align}
    \tilde{P}_{\theta}(y_k|Q) &= \mathbb{E}_{\br{x}} \big{\{} \tilde{P}_{\theta}(y_k|Q, \br{x}) \big{\}} \\
           &\approx \frac{1}{N} \sum_{j=1}^N \tilde{P}_{\theta}(y_k|Q, \br{x}^{(j)}) \\
           &= \frac{1}{N} \sum_{j=1}^N \frac{P_{\theta}(w_k|\br{p}(\br{x}^{(j)}))}{\sum_{w_i} P_{\theta}(w_i|\br{p}(\br{x}^{(j)}))}
\end{align}

\noindent This can be compared to the actual prior of the task/domain,
\begin{align}
    P(y_k) \approx P(y_k|\mathcal{D}) &= \frac{1}{N} \sum_{j=1}^N \mathbbm{1} (y, y^{(j)})
\end{align}

\noindent If $\mathcal{D}$ is sufficiently large, then an unbiased classifier should have a class prior similar to that approximated via the labels. However, if they diverge, one may wish to debias the classifier by scaling class probabilities by factors $\alpha_k$,

\begin{align}
    \hat{P}_{\theta}(y_k| \br{x}, Q, \bm{\alpha}) &= \frac{\alpha_k \tilde{P}_{\theta}(y_k| \br{x}, Q)}{\sum_i \alpha_i \tilde{P}_{\theta}(y_i| \br{x}, Q)}
    \\
    &= \frac{\alpha_k P_{\theta}(w_k| \br{x}, Q)}{Z(\br{x}, Q, \bm{\alpha})}
\end{align}

\noindent Where $Z(\br{x}, Q, \bm{\alpha})= \sum_i \alpha_i P_{\theta}(w_i| \br{x}, Q, \bm{\alpha})$ and $P_{\theta}(w_k|\br{x}, Q) \equiv P_{\theta}(w_k|\br{p}(\br{x}))$. The parameters $\bar{\bm{\alpha}}$ that lead to an unbiased classifier can then be determined in a deterministic fashion. 
\begin{equation}
    \bar{\bm{\alpha}} = \argmin{\bm{\alpha}} \sum_{\forall y_k} | \hat{P}_{\theta}(y_k|Q, \bm{\alpha}) - P(y_k)|
\end{equation}

\noindent Note that by constraining $\alpha_1=1$, there will exist a deterministic solution that ensures that $\hat{P}_{\theta}(y_k|Q, \bm{\alpha}) = P(y_k)$. For given weight parameters $\bm{\alpha}$, Consider the prompt-classifier priors, $\hat{P}_{\theta}(y_k| Q, \bm{\alpha})$. One can approximate this using a Taylor series of the expectation of a ratio, yielding
\begin{align}
    \hat{P}_{\theta}(y_k| Q, \bm{\alpha}) &= \mathbb{E}_{\br{x}}[\hat{P}_{\theta}(y_k| \br{x}, Q, \bm{\alpha})]  \\
                    &= \mathbb{E}_{\br{x}}[\frac{\alpha_k P_{\theta}(w_k| \br{x}, Q)}{Z(\br{x}, Q, \bm{\alpha})}] \\
                    &\approx \frac{\mathbb{E}_{\br{x}}[\alpha_k P_{\theta}(w_k| \br{x}, Q)]}{\mathbb{E}_{\br{x}}[Z(\br{x}, Q, \bm{\alpha})]} \\
                    &= \frac{\alpha_k P_{\theta}(w_k| Q)}{Z(Q, \bm{\alpha})}
\end{align}

\noindent By equating the predicted prior with the true prior, we find an approximation for $\bar{\bm{\alpha_k}}$
\begin{align}
    \hat{P}_{\theta}(y_k| Q) &= P(y_k|\mathcal{D}) \\
    \frac{\alpha_k P_{\theta}(w_k| Q)}{Z(Q)} &= P(y_k|\mathcal{D}) \\
    \Rightarrow \alpha_k &= \frac{Z(Q) \cdot P(y_k|\mathcal{D})}{P_{\theta}(w_k| Q)}
\end{align}

\noindent A final insight is that in many cases it is assumed that there should be no inherent class bias, and so $P(y_k|\mathcal{D})$ can be assumed to be uniform and be included in the normalisation term.

\section{Prompts and Label Words}
\label{sec:example_prompts}
\subsection{Sentiment Classification}
\begin{table}[H]
    \small
    \centering
    \begin{tabular}{l}
        \toprule
        \multicolumn{1}{c}{prompt}           \\
        \midrule
        classify the following review:       \\
        how was the movie?                   \\
        which word best describes the text?  \\
        what is the sentiment?               \\
        what is the reviewer's verdict?      \\
        is the following movie good or bad?  \\
        \bottomrule
    \end{tabular}
    \caption{sentiment classification prompts}
    \label{tab:temp}
\end{table}

\begin{table}[H]
    \small
    \centering
    \begin{tabular}{l|l}
        \toprule
        positive  & negative    \\
        \midrule
        good         & bad        \\
        great        & terrible   \\
        amazing      & poor       \\
        fantastic    & horrible   \\
        positive     & negative   \\
        \bottomrule
    \end{tabular}
    \caption{label words for sentiment classification}
    \label{tab:temp}
\end{table}

\subsection{Natural Language Inference}
\begin{table}[H]
    \small
    \centering
    \begin{tabular}{l}
        \toprule
        \multicolumn{1}{c}{prompt}                             \\
        \midrule
        is the second text an entailment of the first text?        \\
        does the second text directly follow from the first text?  \\
        are the texts related?                                     \\
        are the texts consistent?                                  \\
        does text 1 imply text 2?                                  \\
        can text 2 be logically derived from text 1?               \\
        does the hypothesis logically follow the premise?          \\
        \bottomrule
    \end{tabular}
    \caption{NLI prompts}
    \label{tab:temp}
\end{table}

\begin{table}[H]
    \small
    \centering
    \begin{tabular}{l|l|l}
        \toprule
        entailment  & neutral & contradiction     \\
        \midrule
        yes         & maybe       & no            \\
        correct     & unclear     & incorrect     \\
        yeah        & potentially & nope          \\
        follows     & neutral     & contradiction \\
        \bottomrule
    \end{tabular}
    \caption{label words for NLI}
    \label{tab:temp}
\end{table}

\subsection{Paraphrase Identification}
\begin{table}[H]
    \small
    \centering
    \begin{tabular}{l}
        \toprule
        \multicolumn{1}{c}{prompt} \\
        \midrule
        is the second text a paraphrase of the first text?        \\
        are the two texts semantically equivalent?  \\
        are the texts paraphrases of each other? \\
        do the texts have the same meaning? \\
        is the meaning of text 1 the same as in text 2? \\
        would the two texts be classified as paraphrases?  \\
        \bottomrule
    \end{tabular}
    \caption{NLI prompts}
    \label{tab:temp}
\end{table}

\begin{table}[H]
    \small
    \centering
    \begin{tabular}{l|l}
        \toprule
        paraphrase  & not paraphrase    \\
        \midrule
        yes       & no        \\
        correct   & incorrect   \\
        yeah      & not       \\
        positive  & negative   \\
        true      & false   \\
        \bottomrule
    \end{tabular}
    \caption{label words for sentiment classification}
    \label{tab:temp}
\end{table}

\section{Threshold Alignment Plots}
\label{sec:linear_plots}
\begin{figure}[h!]
    \centering
    \includegraphics[width=0.8\columnwidth]{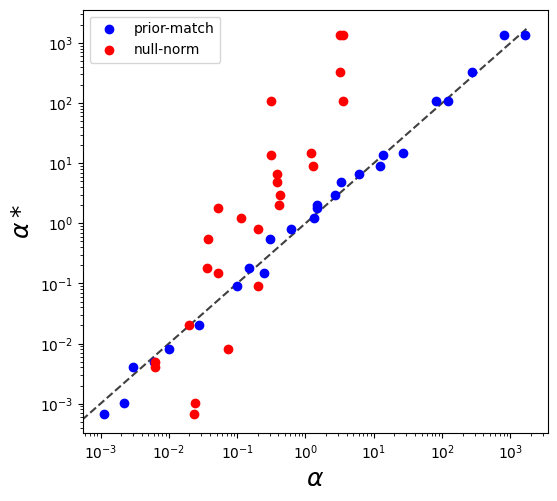}
    \caption{weights alignment plot for \textbf{rotten tomatoes}}
    \label{fig:boxplot_imdb}
\end{figure}

\begin{figure}[h!]
    \centering
    \includegraphics[width=0.8\columnwidth]{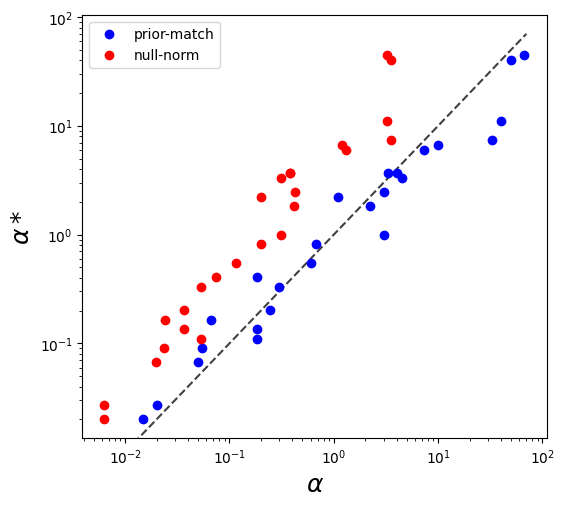}
    \caption{weights alignment plot for \textbf{imdb}}
    \label{fig:boxplot_imdb}
\end{figure}

\onecolumn
\section{Impact of LLM Choice}
\label{sec:app_tables}
\begin{table}[h!]
    \centering
    \begin{tabular}{ccc|cccccc}
        \toprule
        method  & inputs & labels & imdb & rt & amazon & snli & mnli & qqp \\
        \midrule
        baseline       & \xmark & \xmark & 82.1\std{11.1} & 70.1\std{11.7}& 83.4\std{12.8}& 37.4\std{6.1} & 37.0\std{4.3} & 52.5\std{11.4} \\ 
        null-input & \xmark & \xmark      & 87.5\std{3.2}  & 78.5\std{4.8} & 91.1\std{2.3} & 41.8\std{3.8} & 40.2\std{4.1} & 53.9\std{10.2} \\
        prior-match & \cmark & \xmark    & 89.1\std{2.4}  & 80.8\std{2.9} & 92.0\std{1.3} & 44.7\std{6.3} & 41.8\std{3.8} & 58.5\std{5.5} \\
        optimal& \cmark & \cmark         & 89.3\std{2.0}  & 81.2\std{2.9} & 92.1\std{1.4} & 47.6\std{5.9} & 43.5\std{3.7} & 65.3\std{2.9} \\
        \bottomrule
    \end{tabular}
    \caption{Robustness performance when using FlanT5 base as the base LLM (set-up equivalent to Table 1).}
    \label{tab:flan_small_table}
\end{table}

\begin{table}[h!]
    \centering
    \begin{tabular}{ccc|cccccc}
        \toprule
        method      & inputs & labels  & imdb          & rt            & amazon        & snli          & mnli          & qqp \\
        \midrule
        baseline    & \xmark & \xmark  & 85.8\std{8.7} & 78.4\std{10.3}& 86.4\std{10.3}& 35.1\std{2.7} & 36.9\std{3.2} & 51.0\std{11.6} \\ 
        null-input   & \xmark & \xmark  & 87.4\std{6.9} & 83.2\std{6.6} & 90.7\std{6.4} & 37.9\std{5.4} & 39.4\std{3.7} & 51.8\std{8.4} \\
        prior-match & \cmark & \xmark  & 90.5\std{3.1} & 86.3\std{3.7} & 93.1\std{2.4} & 39.5\std{5.4} & 41.2\std{3.1} & 52.6\std{1.9} \\
        optimal     & \cmark & \cmark  & 90.8\std{2.8} & 86.7\std{3.6} & 93.2\std{2.4} & 42.8\std{4.6} & 42.8\std{2.3} & 66.8\std{0.4} \\
        \bottomrule
    \end{tabular}
    \caption{Robustness performance when using Llama-2-chat 7B as the base LLM.}
    \label{tab:llama_table}
\end{table}
 
\noindent Tables \ref{tab:flan_small_table} and \ref{tab:llama_table} show the prompt-based classifier performance for the different methods when using different FlanT5-base and LLama-2-chat 7B respectively. For sentiment classification and natural language tasks, we similarly observe that the various re-weighting methods lead to considerable boosts in accuracy. Both null-norm and prior-match again lead to performance near that to the optimal weights, with considerable performance boost over the baseline. However, for paraphrase detection, we only observe moderate performance boosts over the baseline setting with a larger performance discrepancy with the optimal weights. 

\vspace{3cm}

\section{Boxplots}
\begin{figure}[ht!]
    \centering
    \includegraphics[width=\linewidth]{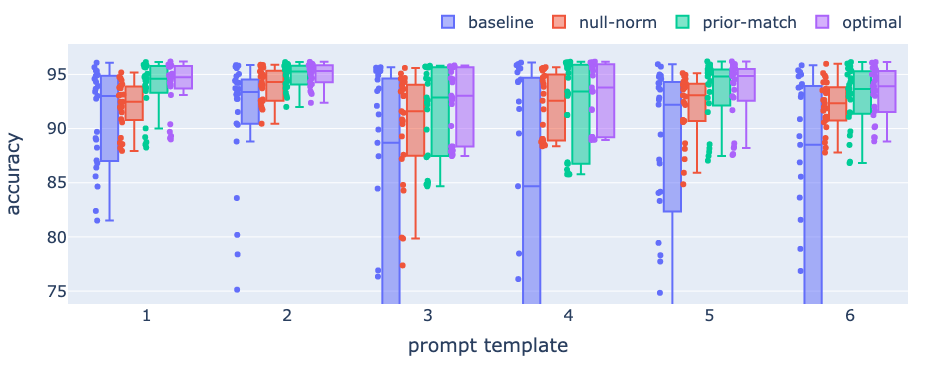}
    \caption{boxplots of the accuracy of all label-word pairs on \textbf{IMDB}, over all the considered prompts}
    \label{fig:none}
\end{figure}

\begin{figure}[ht!]
    \centering
    \includegraphics[width=\linewidth]{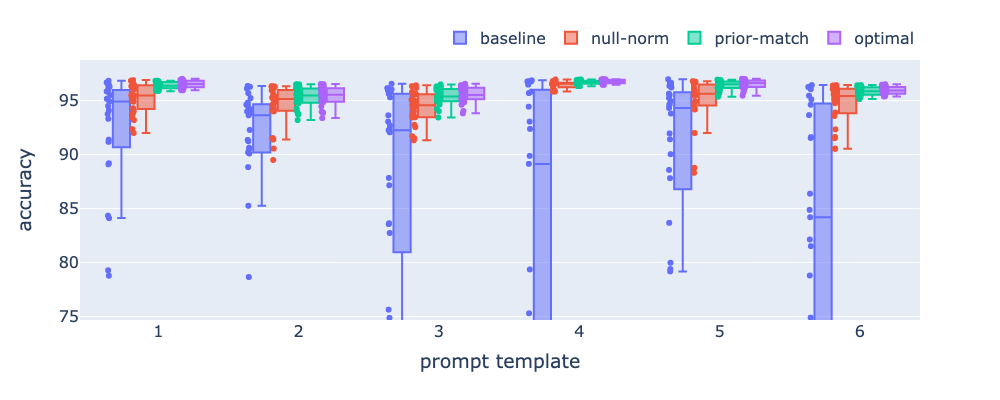}
    \caption{boxplots of the accuracy of all label-word pairs on \textbf{amazon}, over all the considered prompts}    \label{fig:none}
\end{figure}

\begin{figure}[ht!]
    \centering
    \includegraphics[width=\linewidth]{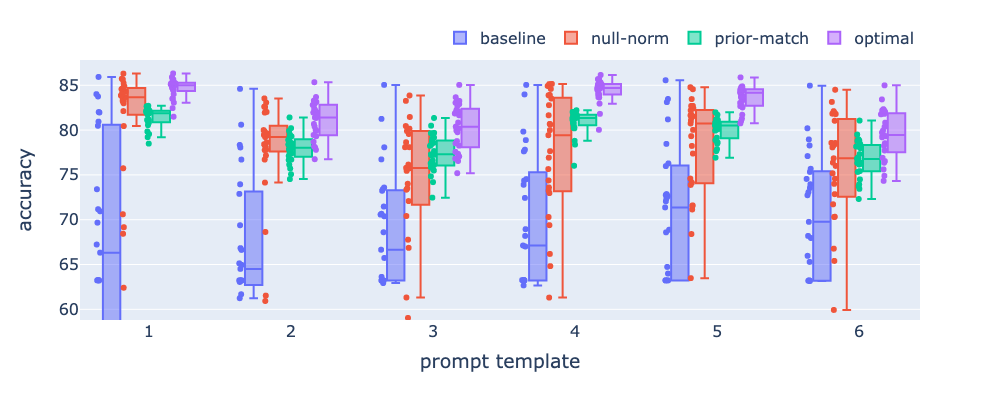}
    \caption{boxplots of the accuracy of all label-word pairs on \textbf{qqp}, over all the considered prompts}    \label{fig:none}
\end{figure}

\vspace{15cm}
\phantom{}

\end{document}